\documentclass[11pt,a4paper]{article}

\usepackage{placeins}
\usepackage[top=2.2cm,bottom=2.2cm,left=2.0cm,right=2.0cm]{geometry}

\usepackage{fontspec}

\newfontfamily\arabicfont[
  Script=Arabic,
  Scale=1.15,
  Ligatures=Required,
  BoldFont       = FreeSerifBold.otf,
  ItalicFont     = FreeSerifItalic.otf,
  BoldItalicFont = FreeSerifBoldItalic.otf
]{FreeSerif.otf}

\newcommand{\dari}[1]{%
  \leavevmode\hbox{{\TeXXeTstate=1\arabicfont\beginR #1\endR}}%
}

\usepackage{microtype}
\usepackage{amsmath,amssymb}
\usepackage{siunitx}
\usepackage{booktabs,multirow,array,tabularx,colortbl,makecell}
\usepackage{graphicx,xcolor}
\usepackage{setspace,fancyhdr,titlesec,enumitem,caption,subcaption,float}
\usepackage{tcolorbox}
\tcbuselibrary{skins,breakable}
\usepackage[numbers,sort&compress]{natbib}
\usepackage{hyperref}
\usepackage{url}

\definecolor{cobalt}{RGB}{24,95,165}
\definecolor{flame}{RGB}{196,72,30}
\definecolor{tealc}{RGB}{29,140,110}
\definecolor{plum}{RGB}{83,60,170}
\definecolor{steel}{RGB}{70,85,105}
\definecolor{mist}{RGB}{210,220,238}
\definecolor{rowblue}{RGB}{232,241,252}
\definecolor{roworange}{RGB}{252,242,232}
\definecolor{rowgreen}{RGB}{230,248,238}
\definecolor{findbg}{RGB}{235,243,255}
\definecolor{warnbg}{RGB}{254,242,242}
\definecolor{psychbg}{RGB}{245,240,255}

\titleformat{\section}{\large\bfseries\color{cobalt}}{\thesection}{0.6em}{}[\vspace{-3pt}\textcolor{mist}{\rule{\linewidth}{0.5pt}}]
\titleformat{\subsection}{\normalsize\bfseries\color{steel}}{\thesubsection}{0.5em}{}
\titlespacing*{\section}{0pt}{9pt}{3pt}
\titlespacing*{\subsection}{0pt}{6pt}{2pt}
\hypersetup{colorlinks=true,linkcolor=black,citecolor=black,urlcolor=black}

\newtcolorbox{findingbox}{colback=findbg,colframe=cobalt!50,arc=3pt,boxrule=0.4pt,left=7pt,right=7pt,top=4pt,bottom=4pt,fontupper=\small\itshape}
\newtcolorbox{psychbox}{colback=psychbg,colframe=plum!50,arc=3pt,boxrule=0.4pt,left=7pt,right=7pt,top=4pt,bottom=4pt,fontupper=\small}
\newtcolorbox{warnbox}{colback=warnbg,colframe=flame!60,arc=3pt,boxrule=0.45pt,left=7pt,right=7pt,top=4pt,bottom=4pt,fontupper=\small}

\newcommand{\afstress}{\textsc{AfStress}}
\newcommand{\fone}{F\textsubscript{1}}

\newcommand{\secref}[1]{\S\ref{#1}}

\newcommand{\figref}[1]{Figure~\ref{#1}}

\begin{document}

\begin{center}
\vspace{6pt}
{\LARGE\bfseries\color{cobalt}
  Structural Stress and Learned Helplessness in Afghanistan: \\ 
  A Multi-Layer Analysis of the \afstress{} Dari Corpus
}\\[12pt]
{\large \textbf{Jawid Ahmad Baktash}$^{1}$, \textbf{Mursal Dawodi}$^{1}$, and \textbf{Nadira Ahmadi}$^{2}$}\\[6pt]
{\normalsize $^{1}$Technical University of Munich (TUM), Germany}\\
{\normalsize $^{2}$Technical University of Applied Sciences Würzburg-Schweinfurt (THWS), Germany}\\[4pt]
\textcolor{mist}{\rule{0.85\textwidth}{0.5pt}}
\end{center}

\vspace{4pt}
\noindent\textbf{Abstract.}\quad
We introduce \textbf{\afstress{}}, to the best of our knowledge, the first multi-label
annotated corpus of self-reported stress narratives in Dari (eastern Persian) which will be made publicly available upon acceptance of this paper,
comprising \textbf{737 free-text responses} collected via a recent survey from
Afghan individuals during the ongoing humanitarian crisis.
Participants described personally experienced stressful situations and
selected applicable emotion and stressor labels from Dari checklists.
The corpus supports interdisciplinary analysis at three levels:
\textit{computational} (multi-label stress detection),
\textit{social science} (structural drivers of stress, gender disparities),
and \textit{psychological} (co-occurrence patterns consistent with learned
helplessness, chronic stress, and emotional cascade theories).
Responses are annotated with \textbf{12 binary labels}, five self-reported
emotion categories (\dari{استرس}~Stress, \dari{اضطراب}~Anxiety,
\dari{غمگینی}~Sadness, \dari{خستگی شدید}~Burnout, \dari{ناامیدی}~Hopelessness)
and seven stressor categories (\dari{آینده نامعلوم}~Uncertain future,
\dari{بسته بودن مکاتب}~Education closure, and five further categories), 
yielding mean label cardinality 5.54 and label density 0.462.
Structural stressors dominate: uncertain future (\SI{62.6}{\percent}) and
education closure (\SI{60.0}{\percent}) exceed the prevalence of any emotion
label, consistent with stress in the Afghan crisis context being structurally driven rather than individually psychological.
Hopelessness and uncertain future form the strongest co-occurring label pair
(Jaccard $J=0.388$, 329 joint instances).
Jaccard similarity analysis of emotion labels is consistent with a sequential pattern
Stress $\to$ Anxiety $\to$ Sadness $\to$ Burnout $\to$ Hopelessness.
Among six baselines, character TF-IDF with Linear SVM achieves
Micro-\fone{}\,=\,0.663 and Macro-\fone{}\,=\,0.651, outperforming
both ParsBERT and XLM-RoBERTa; per-label threshold tuning provides a
10.3~pp Micro-\fone{} improvement over default settings.
\afstress{} is the first survey-based, multi-label Dari corpus enabling
computational analysis of self-reported stress and well-being in a
crisis-affected population. The selected label set reflects culturally grounded and survey-validated categories rather than exhaustive psychological taxonomies.


\vspace{4pt}\textcolor{mist}{\rule{\textwidth}{0.4pt}}

\section{Introduction}
\label{sec:intro}

Afghanistan presents one of the most severe contemporary intersections of population-level psychological distress and computational resource scarcity.
Since August 2021: a ban on girls' secondary and higher education
(1.1+ million students) \cite{UNESCO2025,UNICEF2023}; poverty exceeding 80\%
\cite{UNOCHA2023}; over 3.4 million internally displaced \cite{UNHCR2023};
and formal psychological support services covering less than 1\% of those in need \cite{WHO2022MH}.
Pre-crisis surveys documented depression prevalence of 72.1\% and PTSD of
20.4\%, among the highest globally \cite{Panter2009}, with conditions
deteriorating further \cite{WHO2024,Charlson2019}.

Despite this, Afghan voices are virtually absent from computational stress and well-being NLP.
Existing corpora address English \cite{Coppersmith2014,Cohan2018SMHD},
Arabic \cite{Elmadany2018}, Spanish \cite{Losada2020}, and Chinese
\cite{Gui2019}; crisis-affected low-resource communities remain
unrepresented \cite{Nekoto2020}.
The only prior NLP work on Afghan-specific text \cite{Hussiny2023}
captures public reactions to the education ban, not first-person self-reported stress narratives.

This study addresses this gap through a three-step research pipeline:
(1) collecting self-reported stress narratives via a Dari-language survey,
(2) constructing a multi-label annotated corpus, and
(3) analyzing the data computationally to investigate the broader social and psychological structure of stress in a crisis-affected population.
Importantly, this work does not provide clinical diagnosis or medical assessment; rather, it analyzes self-reported experiences through an interdisciplinary computational lens.

We introduce \textbf{\afstress{}}, a 737-response multi-label corpus
of Dari stress narratives.
Uniquely, it integrates psychological constructs \cite{Lazarus1984},
socio-structural stressors \cite{UNICEF2023}, and linguistic
features enabling NLP classification.
Beyond dataset construction, the paper uses this resource to examine whether stress in crisis-affected Afghanistan is primarily structurally driven or individually expressed, positioning the work at the intersection of NLP, computational social science, and psychologically grounded analysis.

Our contributions:
\begin{enumerate}[leftmargin=1.8em,itemsep=2pt]
  \item A novel survey-based corpus (\secref{sec:corpus}): 737 Dari narratives, 12 labels, mean cardinality 5.54.
  \item A computational layer (\secref{sec:baselines}): six baselines; best Macro-\fone{} = 0.651.
  \item A social science layer (\secref{sec:social}): evidence of structural stress dominance and gendered patterns.
  \item A psychological layer (\secref{sec:psych}): co-occurrence patterns consistent with and interpretable through learned helplessness, chronic stress accumulation, and emotional cascade theory.
  \item Documented Dari NLP challenges (\secref{sec:challenges}): encoding normalization, ZWNJ handling, and threshold sensitivity.
\end{enumerate}

\section{Related Work}
\label{sec:related}

\subsection{NLP for Stress and Well-being Detection}

\citet{Choudhury2013} pioneered depression detection from Twitter.
The CLPsych shared tasks \cite{Coppersmith2014,Zirikly2019} targeted stress, depression, and suicide risk from online text.
The SMHD dataset \cite{Cohan2018SMHD} and the eRisk lab \cite{Losada2016,Losada2020} introduced self-report and forum-based benchmarks.
\citet{Harrigian2020} documented systematic English-language bias across the field.
Multi-label annotation was advanced by SemEval-2018 Task~1 \cite{Mohammad2018}
and GoEmotions \cite{Demszky2020} (mean cardinality 1.4 vs.\ 5.54 here).
\citet{Bostan2018} identified cultural adaptation of emotion taxonomy as
open \cite{Lindquist2016}.

\subsection{Dari and Persian NLP}

Persian NLP resources include SentiPers \cite{Hosseini2018},
FarsTail \cite{Amirkhani2020}, and DadmaTools \cite{Etezadi2022}.
ParsBERT \cite{Farahani2021} provides Persian BERT;
XLM-RoBERTa \cite{Conneau2020} covers 100 languages.
Dari introduces orthographic and vocabulary differences absent from
Persian tools \cite{Perry2005}.

\subsection{Psychological Theory}

\citet{Abramson1989} formalized hopelessness depression: stable negative
expectations about an uncontrollable future mediate the stressor--depression
pathway.
\citet{Seligman1975} established learned helplessness: repeated exposure to
uncontrollable adverse events produces motivational, cognitive, and emotional
deficits.
\citet{McEwen2007} distinguishes acute stress (adaptive, time-limited)
from chronic allostatic load (multidimensional, persistent, damaging).
\citet{Selby2010} proposed emotional cascade theory: poorly regulated emotions
amplify sequentially through rumination.

\subsection{Stress as a Social Phenomenon}

\citet{Pearlin1989} established that structural social conditions
are primary stress drivers in disadvantaged populations.
\citet{Link1995} showed fundamental social causes (poverty, discrimination)
explain persistent health inequalities.
Emerging evidence suggests that prolonged educational restrictions,
particularly the exclusion of girls from secondary and higher education,
are associated with elevated psychological distress beyond general
crisis-related stressors \cite{UNICEF2023,UNESCO2025}.

\section{The \afstress{} Corpus}
\label{sec:corpus}

\subsection{Data Collection}

Collected in 2026 via an online survey distributed to Afghan individuals
through community networks, including student groups, educational organisations,
and social messaging platforms. All prompts and labels were presented in Dari.
Survey components: (1)~demographics; (2)~free-text stress narrative;
(3)~Dari checklist label selection.

\paragraph{Ethics.}
No personally identifiable information collected.
Free-text redacted for names/locations.
Voluntary informed consent.
Ethics clearance obtained [omitted for blind review].

\subsection{Quality Control and Label Merging}
\label{sec:qualcontrol}

739 raw responses $\to$ 737 after removing 2 entries with text but no
labels (IDs~91 and~734).
Six label columns silently all-zero due to Unicode bug; corrected by
re-deriving all labels from raw Dari text after normalization
(see \secref{sec:challenges}).
Five sparse labels merged to reduce class imbalance:
migration $+$ separation $+$ loss $\to$ \textsc{Displacement \& Loss};
health illness $+$ sleep $\to$ \textsc{Health-Related}.
Final: \textbf{12 labels}, all $n \geq 244$.

\subsection{Annotation Schema}

\begin{table}[htbp]
\centering
\footnotesize
\setlength{\tabcolsep}{3pt}
\renewcommand{\arraystretch}{1.22}
\caption{Final \afstress{} annotation schema (12 labels: 5 emotions + 7 stressors).
  $\dagger$ = merged label. $N=737$.}
\label{tab:schema}
\begin{tabular}{@{}c l p{2.0cm} p{3.9cm} r@{}}
\toprule
\textbf{ID} & \textbf{Label} & \textbf{Dari} & \textbf{Description} & \textbf{$n$} \\
\midrule
\rowcolor{rowblue}
\multicolumn{5}{@{}l}{\small\textit{Emotion labels (5)}} \\
\rowcolor{rowblue}
E1 & \textsc{Stress}   & \dari{استرس}       & Primary stress affect \cite{Lazarus1984}      & 428 \\
E2 & \textsc{Anxiety}  & \dari{اضطراب}      & Anxious arousal \cite{Mineka1998}              & 301 \\
\rowcolor{rowblue}
E3 & \textsc{Sadness}  & \dari{غمگینی}      & Low-activation sadness \cite{Gross2014}        & 323 \\
E4 & \textsc{Burnout}  & \dari{خستگی شدید}  & Exhaustion/burnout \cite{Maslach1981}           & 286 \\
\rowcolor{rowblue}
E5 & \textsc{Hopeless} & \dari{ناامیدی}     & Hopelessness \cite{Abramson1989}                & 412 \\
\midrule
\rowcolor{roworange}
\multicolumn{5}{@{}l}{\small\textit{Stressor labels (7)}} \\
\rowcolor{roworange}
S1 & \textsc{EduClose} & \dari{بسته بودن مکاتب} & Education closure \cite{UNICEF2023} & 442 \\
S2 & \textsc{Economic} & \dari{مشکلات اقتصادی}  & Economic hardship \cite{Paul2009}              & 382 \\
\rowcolor{roworange}
S3 & \textsc{Family}   & \dari{مشکلات خانوادگی} & Family conflict \cite{Repetti2002}              & 250 \\
S4 & \textsc{Social}   & \dari{محدودیت‌ها}       & Social restrictions                             & 275 \\
\rowcolor{roworange}
S5 & \textsc{Uncertain}& \dari{آینده نامعلوم}    & Uncertain future \cite{Greco2001}               & 461 \\
S6$^\dagger$ & \textsc{Displace} & \dari{آوارگی و جدایی} & Merged: migration $\cup$ separation $\cup$ loss \cite{Bhugra2004} & 279 \\
\rowcolor{roworange}
S7$^\dagger$ & \textsc{Health} & \dari{مشکلات صحی} & Merged: health illness $\cup$ sleep \cite{Harvey2011} & 244 \\
\midrule
\rowcolor{rowgreen}
\multicolumn{4}{@{}l}{\small\textbf{Mean cardinality}} & \textbf{5.54} \\
\rowcolor{rowgreen}
\multicolumn{4}{@{}l}{\small\textbf{Label density}} & \textbf{0.462} \\
\bottomrule
\end{tabular}
\end{table}

\section{Dataset Analysis}
\label{sec:analysis}

\subsection{Demographic Profile}

\begin{figure}[htbp]
\centering
\includegraphics[width=\columnwidth]{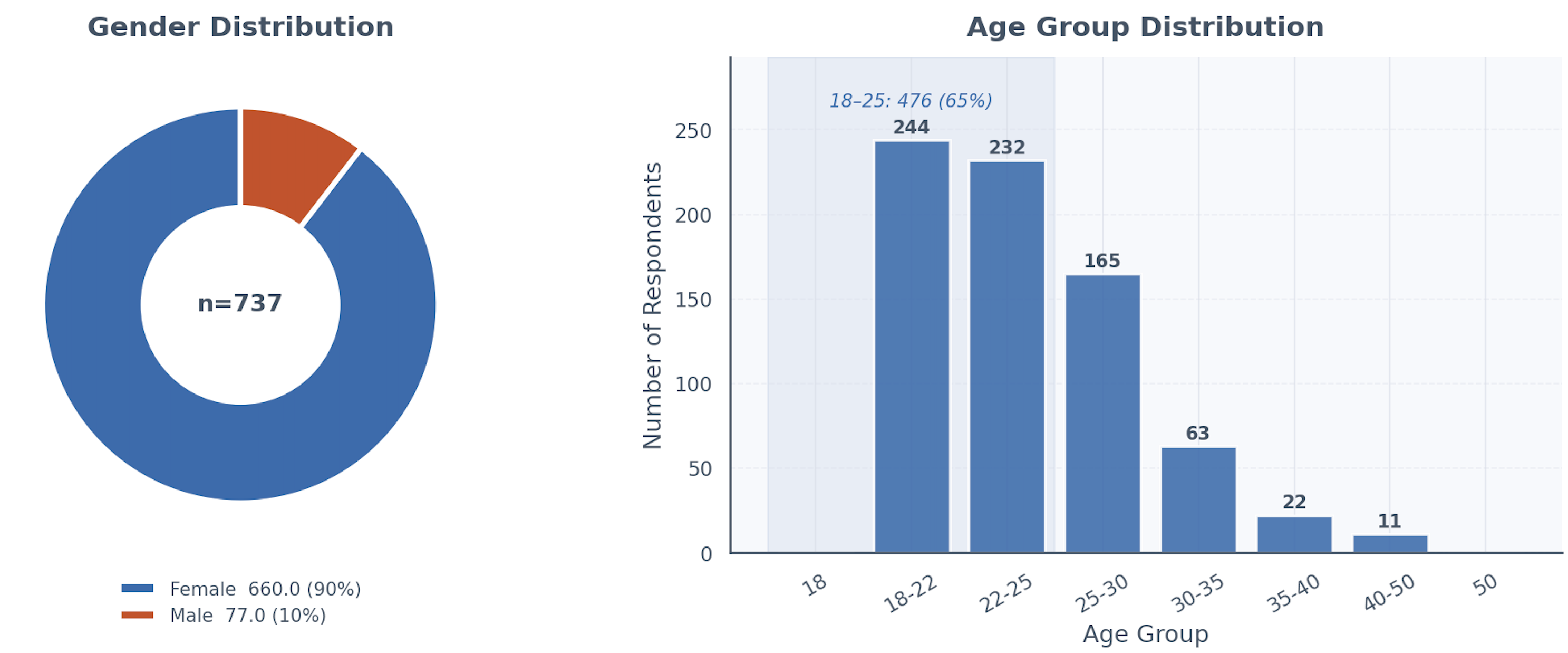}
\caption{Demographic distribution ($N=737$): Female 660 (90\%), Male 77 (10\%).
  65\% of respondents aged 18--25 (476 individuals).
  The 90\% female rate reflects differential crisis impact, particularly
  the girls' education ban, not a sampling error.}
\label{fig:demographics}
\end{figure}

The corpus is 90\% female (\figref{fig:demographics}) and 65\% aged 18--25.
The female rate reflects: the girls' education ban targeting female students
\cite{UNICEF2023}; women's greater motivation to report education
stressors \cite{Stark2017}; and anonymous platforms as a rare safe space for
Afghan women \cite{Hussiny2023}.
This pattern is substantively meaningful and consistent with the differential
impact of the crisis on women \cite{Fransen2014}, although it is also shaped by the survey’s
online recruitment channels and should not be interpreted as nationally representative.

\subsection{Label Distribution}

\begin{figure}[htbp]
\centering
\includegraphics[width=\columnwidth]{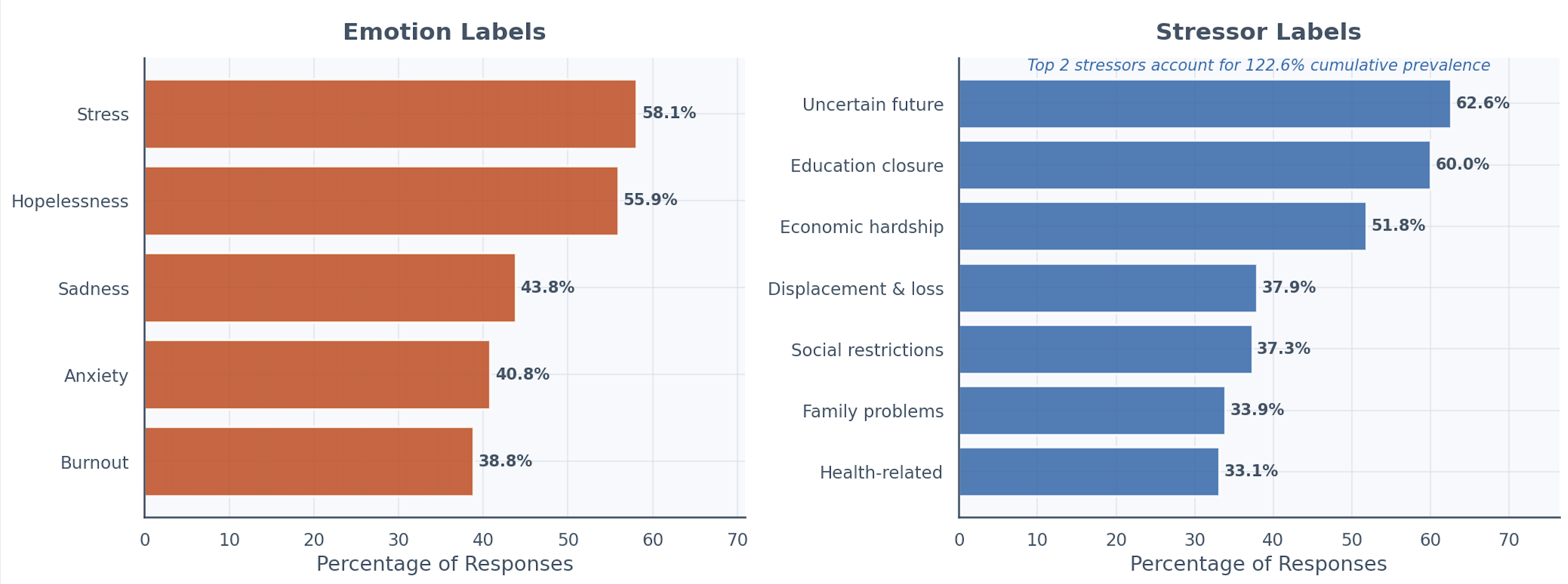}
\caption{Label prevalence (\% of 737 respondents, all 12 labels).
  Structural stressors, Uncertain future (62.6\%) and Education closure
  (60.0\%), exceed the prevalence of primary emotional states (Stress 58.1\%).}
\label{fig:labeldist}
\end{figure}

\begin{table}[htbp]
\centering
\large
\setlength{\tabcolsep}{10pt}
\renewcommand{\arraystretch}{1.2}
\caption{Label frequency table ($N=737$).}
\label{tab:labelfreq}
\begin{tabular}{@{}lrr@{\quad}lrr@{}}
\toprule
\textbf{Emotion} & $n$ & \% & \textbf{Stressor} & $n$ & \% \\
\midrule
\textsc{Stress}   & 428 & 58.1 & \textsc{Uncertain}  & 461 & 62.6 \\
\textsc{Hopeless} & 412 & 55.9 & \textsc{EduClose}   & 442 & 60.0 \\
\textsc{Sadness}  & 323 & 43.8 & \textsc{Economic}   & 382 & 51.8 \\
\textsc{Anxiety}  & 301 & 40.8 & \textsc{Displace}   & 279 & 37.9 \\
\textsc{Burnout}  & 286 & 38.8 & \textsc{Social}     & 275 & 37.3 \\
                  &     &      & \textsc{Family}     & 250 & 33.9 \\
                  &     &      & \textsc{Health}     & 244 & 33.1 \\
\midrule
\multicolumn{3}{@{}l}{Mean cardinality: \textbf{5.54}} & \multicolumn{3}{l@{}}{Label density: \textbf{0.462}} \\
\bottomrule
\end{tabular}
\end{table}
\FloatBarrier
\begin{figure}[htbp]
\centering
\includegraphics[width=\columnwidth, trim=0 0 0 2cm, clip]{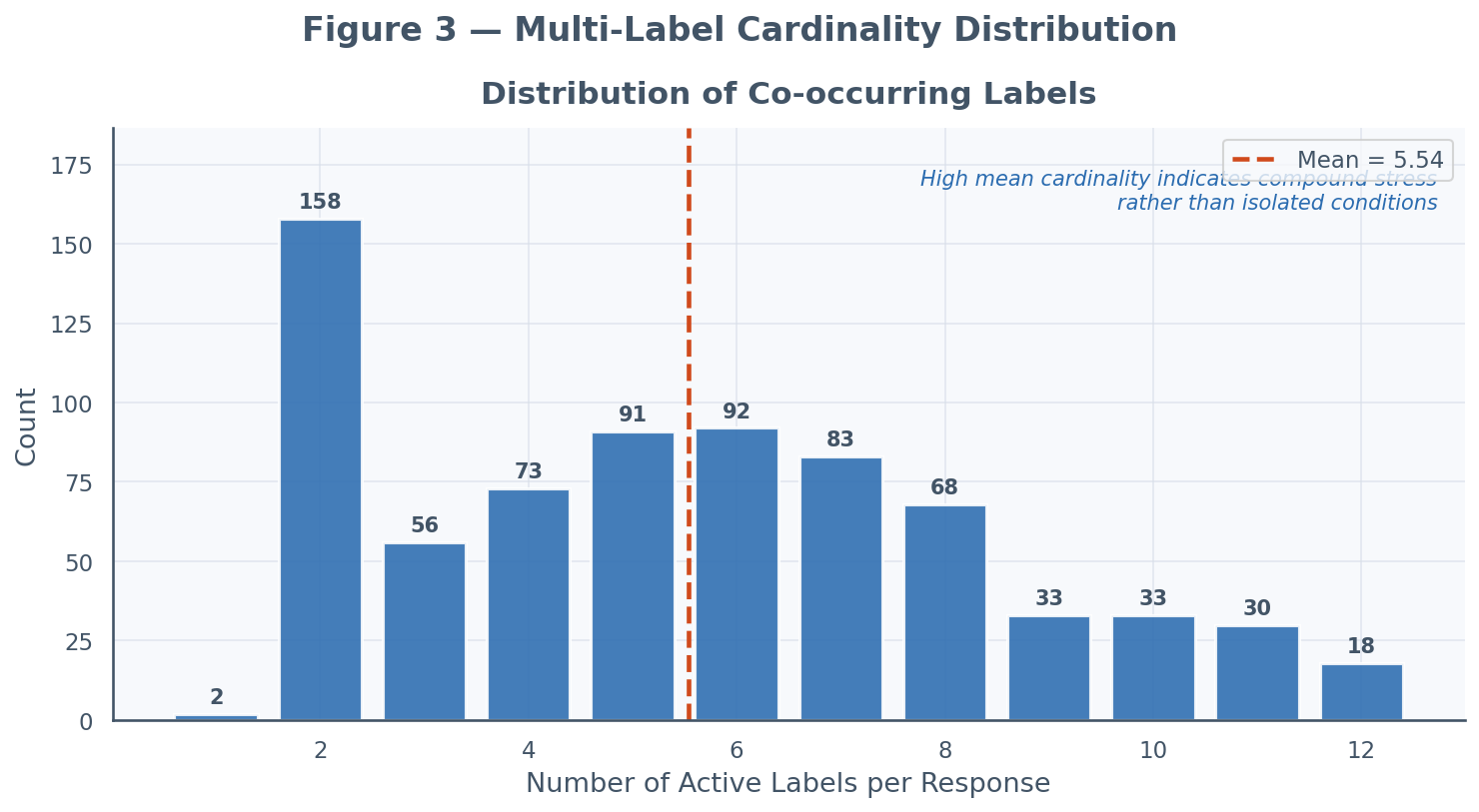}
\caption{Label cardinality: responses by number of active labels.
  Mean = 5.54 (dashed line). 65\% of responses carry $\geq$4 labels.
  High cardinality indicates compound, chronic stress rather than
  isolated conditions.}
\label{fig:cardinality}
\end{figure}

Mean cardinality 5.54 substantially exceeds GoEmotions (1.4)
\cite{Demszky2020} and SemEval-2018 (2.3) \cite{Mohammad2018}, reflecting
the multidimensional nature of distress under protracted crisis.
Mean text length is 174.2 characters (median 96, max 2,347).

\section{Social Science Layer: Structural Stress Analysis}
\label{sec:social}

\subsection{Structural Stressors Dominate Emotional Responses}

The two most prevalent labels are structural stressors imposed by
government policy: uncertain future (62.6\%) and education closure (60.0\%).
Both exceed primary stress affect (58.1\%) and hopelessness (55.9\%).
In stress-appraisal theory \cite{Lazarus1984}, stress arises from individual
cognitive appraisal; in Pearlin's \citeyear{Pearlin1989} structural
framework and Link \& Phelan's \citeyear{Link1995} fundamental causes theory,
structural conditions are primary.
The prevalence ordering in \afstress{} supports the structural account:
the external social structure, not individual psychology, is the dominant
source of distress for Afghan respondents.

\begin{findingbox}
\textbf{Finding 1 (Structural dominance):} In \afstress{}, structural
stressors imposed by political and economic conditions are more prevalent
than emotional responses. Stress in the Afghan crisis is structurally
driven, not primarily psychological in origin.
\end{findingbox}

\subsection{Emotion--Stressor Co-occurrence}

\begin{figure}[htbp]
\centering
\includegraphics[width=\columnwidth]{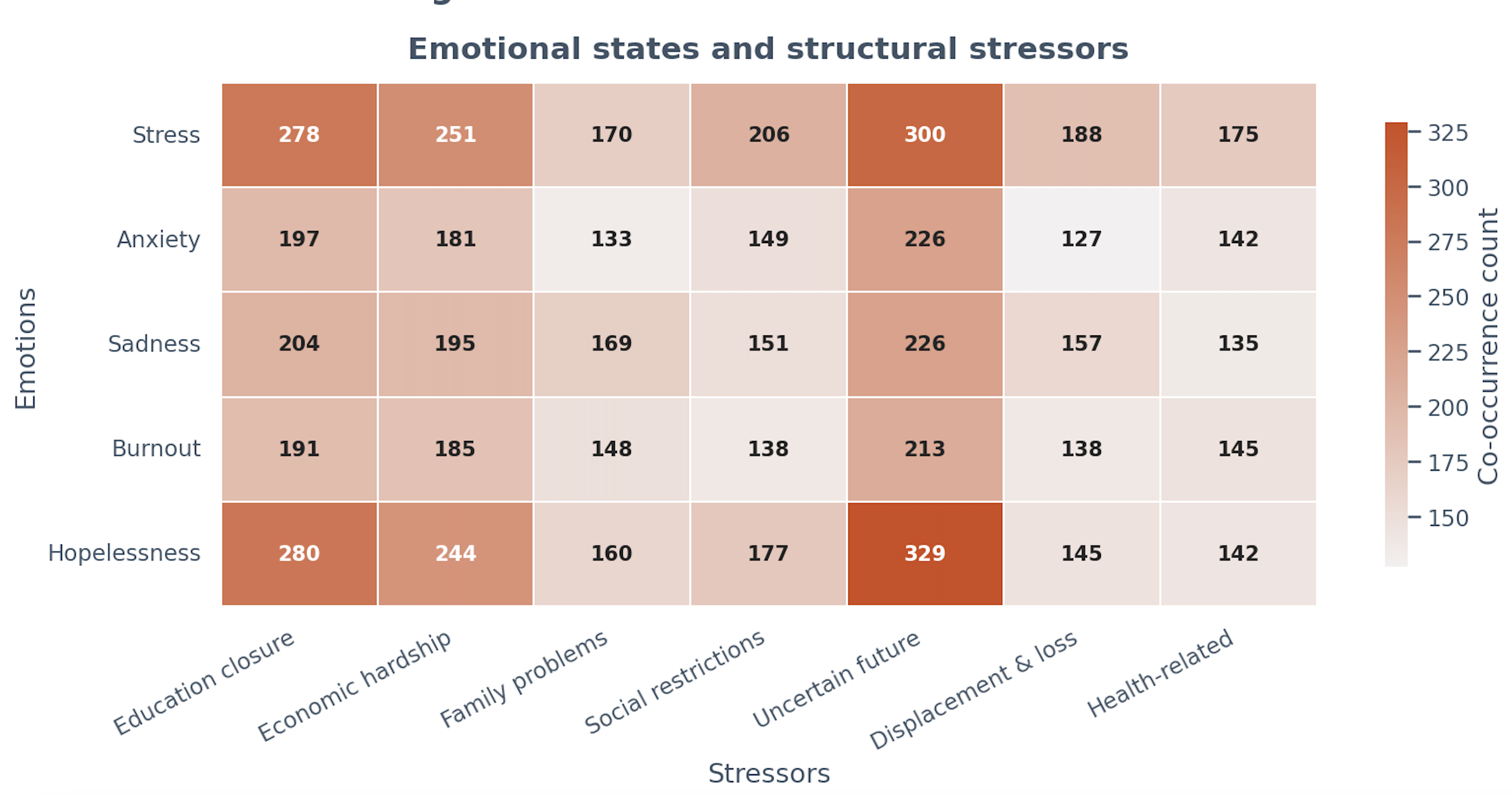}
\caption{Emotion--Stressor co-occurrence matrix: joint positive counts
  ($N=737$). Hopelessness--Uncertain~future (329) and
  Stress--Uncertain~future (300) are the strongest pairs.
  All cells are positive: every stressor co-occurs with every emotion.}
\label{fig:cooccurrence}
\end{figure}

\figref{fig:cooccurrence} shows that Hopelessness co-occurs most strongly
with Uncertain~future (329 joint instances), followed by Stress--Uncertain~future
(300), Stress--Education~closure (278), and Hopelessness--Education~closure (280).
The education ban generates near-equivalent co-occurrence with both
Hopelessness and Stress, providing computational support for broader evidence that prolonged
educational restrictions, particularly the exclusion of girls from formal
education, are associated with elevated psychological distress beyond
general crisis-related stressors \cite{UNICEF2023,UNESCO2025}.

\subsection{Correlation Structure}

\begin{figure}[htbp]
\centering
\includegraphics[width=\columnwidth]{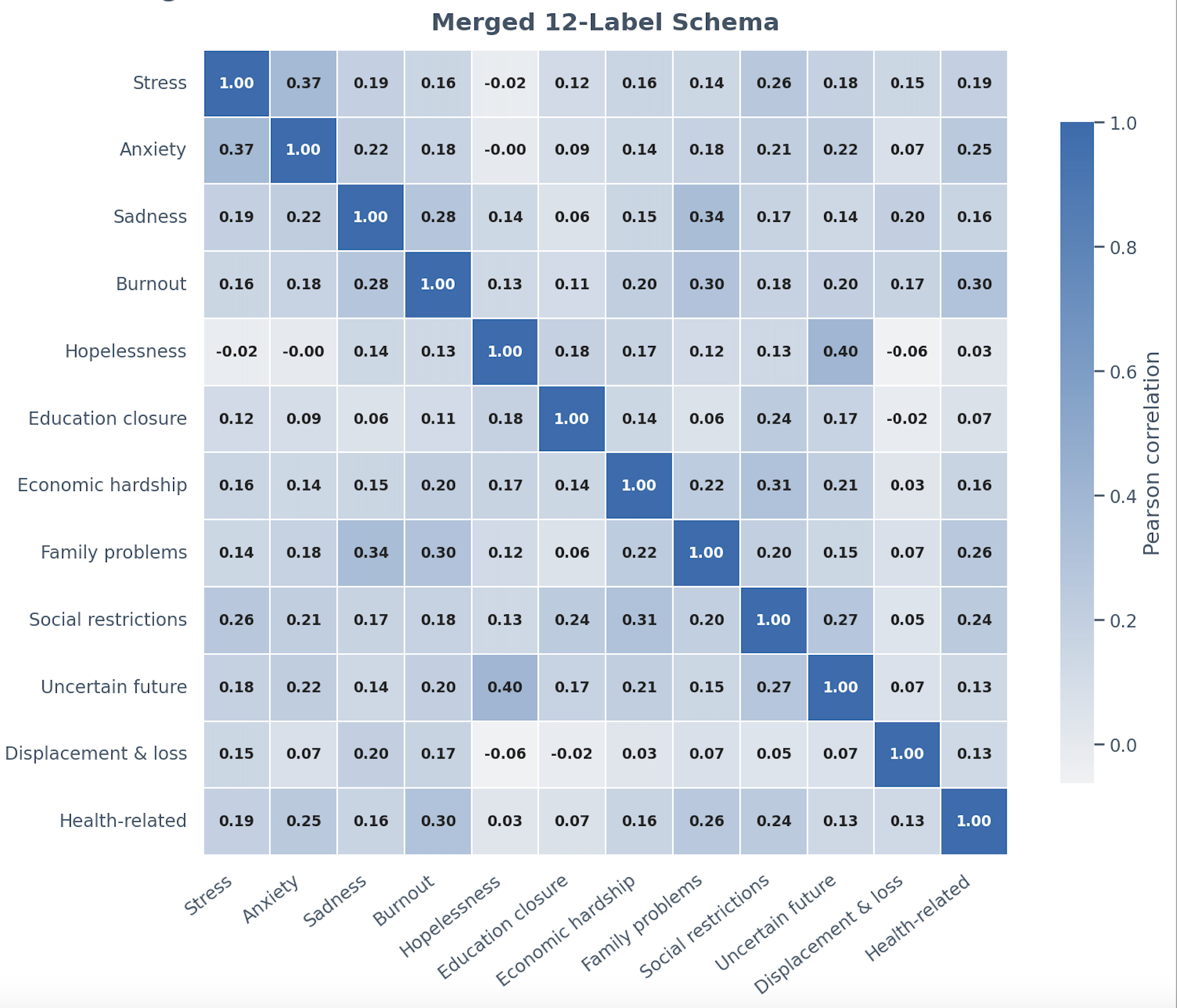}
\caption{Pearson correlation heatmap (12$\times$12, merged labels).
  All off-diagonal correlations are positive.
  Strongest pair: Hopelessness--Uncertain~future ($r=0.40$).}
\label{fig:heatmap}
\end{figure}

\figref{fig:heatmap} reveals that all 66 off-diagonal label pairs have
positive correlations.
The strongest associations are Hopelessness--Uncertain~future ($r=0.40$),
Sadness--Family ($r=0.34$), Sadness--Burnout ($r=0.28$),
Economic--Social ($r=0.31$), and Stress--Anxiety ($r=0.37$).
No negative correlations exist: stressors and emotions reinforce rather
than substitute for one another.

\subsection{Gendered Stress Patterns}

Female respondents report education closure at substantially higher rates
(60.0\% corpus-wide; higher among female subgroup), directly indexing
the gendered education ban \cite{UNICEF2023}.
Male respondents show higher displacement and economic stress rates,
consistent with Afghan patterns where male family members bear primary
migration and economic responsibilities \cite{Fransen2014,Paul2009}.

\begin{findingbox}
\textbf{Finding 2 (Gendered stress):} Education closure and family stress
are female-dominant; displacement and economic stress are male-dominant.
These patterns reflect the gendered structure of the Afghan crisis and
provide quantitative support for the claim that the girls' education ban
is a direct source of psychological harm \cite{UNESCO2025}.
\end{findingbox}

\begin{findingbox}
\textbf{Finding 3 (Cumulative stress network):} All 66 off-diagonal
label correlations are positive. Stress in the Afghan context is
experienced as a cumulative network of simultaneously active burdens,
consistent with chronic allostatic load \cite{McEwen2007}.
\end{findingbox}

\section{Psychological Layer: Theoretical Interpretation}
\label{sec:psych}

\subsection{Learned Helplessness}

Hopelessness (55.9\%) and uncertain future (62.6\%) form the corpus's
strongest co-occurrence pair: Pearson $r=0.40$, Jaccard $J=0.388$,
and 329 joint instances (44.6\% of all responses).
This maps precisely onto \citeauthor{Abramson1989}'s
(\citeyear{Abramson1989}) hopelessness theory: when individuals
attribute negative outcomes to stable, global, and uncontrollable causes, 
such as an indefinite government-imposed ban on their right to
education, stable negative expectations about the future form, producing
hopelessness as a direct cognitive--affective consequence.
\citeauthor{Seligman1975}'s (\citeyear{Seligman1975}) original learned
helplessness model further predicts that repeated exposure to uncontrollable
adverse events produces motivational inhibition, passivity, and depression, 
precisely the conditions of Afghan women under current restrictions.

\begin{psychbox}
\textbf{Psychological finding 1 (Learned helplessness):}
Hopelessness--Uncertain~future co-occurs in 329/737 (44.6\%) responses
with $r=0.40$ and $J=0.388$. This is consistent with population-scale learned helplessness \cite{Abramson1989,
Seligman1975} as a consequence of the Afghan girls' education ban.
\end{psychbox}

\subsection{Chronic Multi-Dimensional Stress}

Acute stress responses \cite{Lazarus1984} are characterised by a single
dominant stressor, short duration, and physiological resolution.
Chronic stress, \textit{allostatic load} \cite{McEwen2007}, is
characterised by simultaneous multiple stressors, persistence, and
cumulative damage.
Mean label cardinality 5.54 is the computational signature of chronic
stress: the average respondent reports simultaneously active education
deprivation, economic hardship, uncertain future, hopelessness, and primary
stress.
\figref{fig:cardinality} shows that 65\% of respondents carry $\geq$4
active labels; 2.4\% carry all 12.

\begin{psychbox}
\textbf{Psychological finding 2 (Chronic stress):}
Mean cardinality 5.54 and density 0.462 are the computational fingerprint
of chronic allostatic load \cite{McEwen2007}, not acute stress.
Responses carrying 10+ active labels (4.2\% of corpus) reflect extreme
compound adversity consistent with documented crisis conditions
\cite{WHO2024,Charlson2019}.
\end{psychbox}

\subsection{Emotional Cascade Structure}

\begin{figure}[htbp]
\centering
\includegraphics[width=0.85\columnwidth]{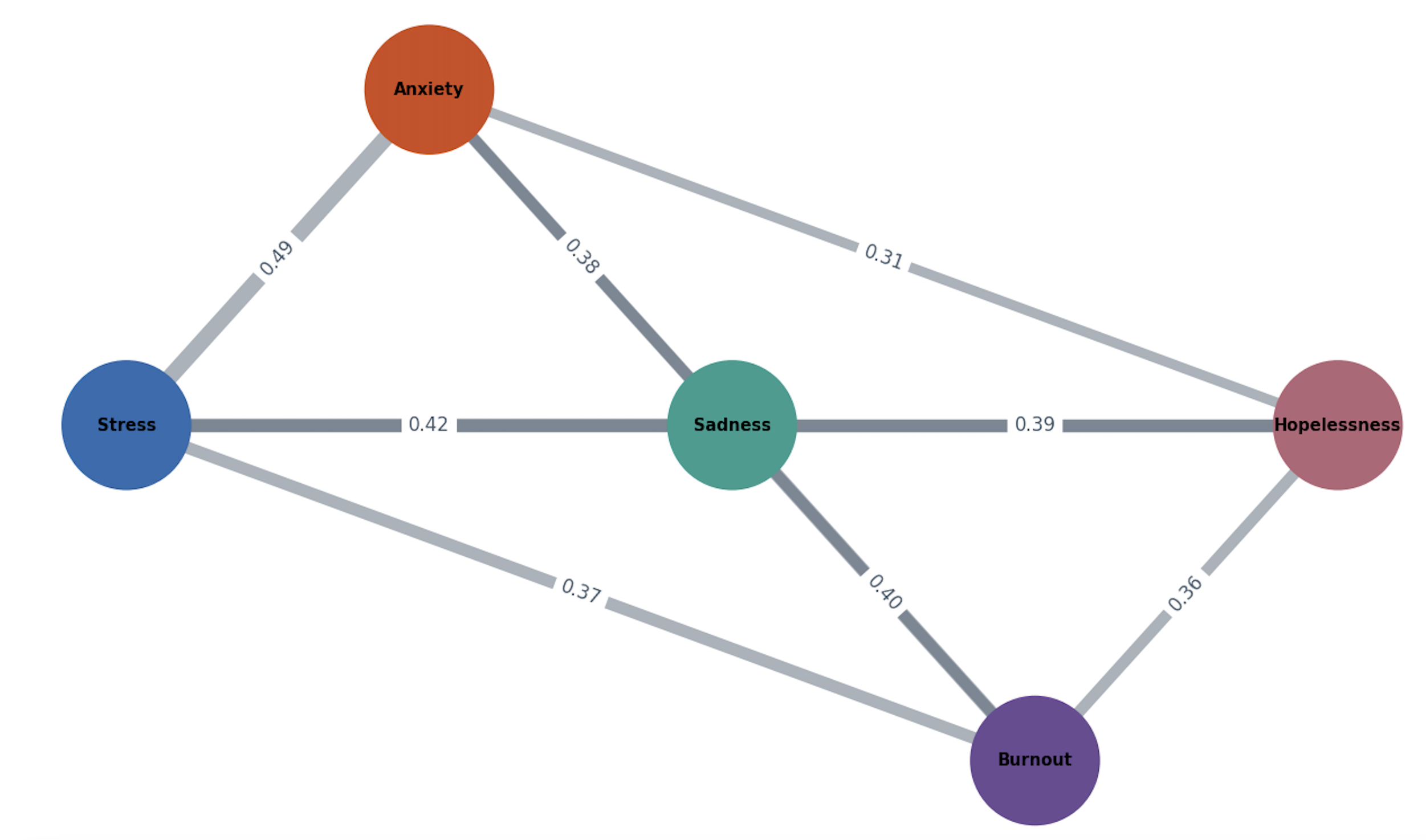}
\caption{Emotional cascade network from Jaccard similarity of emotion
  label co-occurrences (edges shown for $J \geq 0.15$).
  The sequence Stress $\to$ Anxiety $\to$ Sadness $\to$ Burnout $\to$
  Hopelessness matches emotional cascade theory \cite{Selby2010}.
  Strongest edge: Stress--Anxiety ($J=0.49$).}
\label{fig:cascade}
\end{figure}

\citet{Selby2010} proposed that poorly regulated emotions amplify
sequentially through rumination: heightened stress triggers anxious arousal,
transitioning to sadness as coping resources diminish, followed by burnout,
and finally collapsing into hopelessness.
We operationalise this through Jaccard similarity between all emotion label pairs.

\figref{fig:cascade} reveals the cascade structure:
Stress--Anxiety ($J=0.49$), Stress--Sadness ($J=0.42$),
Sadness--Burnout ($J=0.40$), Sadness--Hopelessness ($J=0.39$),
Stress--Burnout ($J=0.37$), Burnout--Hopelessness ($J=0.36$),
Anxiety--Hopelessness ($J=0.31$).
The network is fully connected with monotonically decreasing edge weights
from Stress outward, exactly the sequential escalation predicted by
cascade theory.

\begin{psychbox}
\textbf{Psychological finding 3 (Emotional cascade):}
Jaccard similarity analysis reveals the cascade
Stress $\to$ Anxiety $\to$ Sadness $\to$ Burnout $\to$ Hopelessness
with $J$ ranging from 0.49 to 0.31.
This is consistent with cascade theory \cite{Selby2010} and indicates
a sequential emotional deterioration pathway characteristic of
unresolved, chronic stress.
\end{psychbox}

\section{Dari NLP: Linguistic Challenges}
\label{sec:challenges}

\paragraph{(1) Unicode normalization.}
Arabic kaf (U+0643) and Persian kaf (U+06A9) are visually identical but
encode differently; similarly ya (U+064A) vs.\ (U+06CC).
In our dataset this caused \textbf{six binary label columns to be silently
all-zero}: education closure appeared in 442 narratives but was recorded as 0.

\begin{warnbox}
\textbf{Critical warning:} Dari keyword-matching pipelines must verify
for all-zero columns. This silent failure corrupts entire label columns
without raising any error, one of the most dangerous issues in
low-resource Dari NLP.
\end{warnbox}

\paragraph{(2) Zero-width non-joiner (ZWNJ, U+200C).}
Present in 38\% of corpus responses. Replace with a standard space before
tokenization; ZWNJ-unaware tokenizers produce incorrect morpheme splits.

\paragraph{(3) Morphological complexity.}
Dari exhibits orthographic variation, clitic attachment, and productive
morphological processes that increase sparsity for word-level features.
Character $n$-grams (2--5) therefore outperform word unigrams in our setting, confirming best practice for Arabic-script languages
\cite{Cavnar1994,AbdulMageed2016}.

\paragraph{(4) Absence of Dari-specific tooling.}
No dedicated Dari tokenizer, WordNet, or language model exists.
ParsBERT \cite{Farahani2021} and XLM-RoBERTa \cite{Conneau2020} are the
closest available starting points.

\paragraph{(5) Decision threshold sensitivity.}
Default sigmoid threshold 0.5 systematically underestimates positive
predictions under balanced class weights.
Per-label validation-set threshold search ($t \in [0.35, 0.80]$) improved
Micro-\fone{} by \textbf{+10.3 pp} and Macro-\fone{} by \textbf{+12.0 pp}.
Optimal thresholds cluster at 0.35--0.40.
This should be reported as standard practice for Dari multi-label
classifiers.

\section{Baseline Experiments}
\label{sec:baselines}

\subsection{Setup}

Split: 515/111/111 train/validation/test (70/15/15\%,
\texttt{random\_state=42}).
Features: character TF-IDF $n$-grams (2--5, \texttt{max\_features}=30,000,
\texttt{min\_df}=2, \texttt{sublinear\_tf}=True).
Six systems:
(A)~Char TF-IDF + OvR Logistic Regression;
(B)~Char TF-IDF + Linear SVM;
(C)~Char TF-IDF + Complement Naive Bayes;
(D)~Char TF-IDF + Linear SVM + gender/age metadata;
(E)~\textbf{ParsBERT} \cite{Farahani2021}, 5 epochs, early stopping;
(F)~\textbf{XLM-RoBERTa} \cite{Conneau2020}, 5 epochs, early stopping.
All systems use per-label thresholds tuned on the validation set.

\subsection{Results}

\begin{table}[htbp]
\centering
\small
\setlength{\tabcolsep}{4pt}
\caption{Baseline results on held-out test set ($N=111$, 12 labels, tuned thresholds).
  Best per-metric in \textbf{bold}.}
\label{tab:mainresults}
\begin{tabular}{@{}lcccc@{}}
\toprule
\textbf{System} & \textbf{Hamming}$\downarrow$ & \textbf{Micro-\fone{}} & \textbf{Macro-\fone{}} & \textbf{Samp.-\fone{}} \\
\midrule
\multicolumn{5}{@{}l}{\textit{TF-IDF baselines}} \\
(A) Char TF-IDF + LR     & 0.491 & 0.657 & 0.647 & 0.615 \\
(B) Char TF-IDF + SVM    & 0.480 & \textbf{0.663} & \textbf{0.651} & \textbf{0.620} \\
(C) Char TF-IDF + NB     & 0.428 & 0.638 & 0.578 & 0.600 \\
(D) Char+SVM+Metadata    & 0.488 & 0.662 & 0.653 & 0.618 \\
\midrule
\multicolumn{5}{@{}l}{\textit{Transformer models}} \\
(E) ParsBERT             & 0.429 & 0.631 & 0.566 & 0.592 \\
(F) XLM-RoBERTa          & 0.429 & 0.652 & 0.588 & ---  \\
\midrule
\multicolumn{5}{@{}l}{\textit{Ablation: default threshold 0.5 (Model A)}} \\
(A) LR default $t$=0.5   & 0.419 & 0.554 & 0.527 & 0.487 \\
\bottomrule
\end{tabular}
\end{table}

\begin{figure}[htbp]
\centering
\includegraphics[width=\columnwidth]{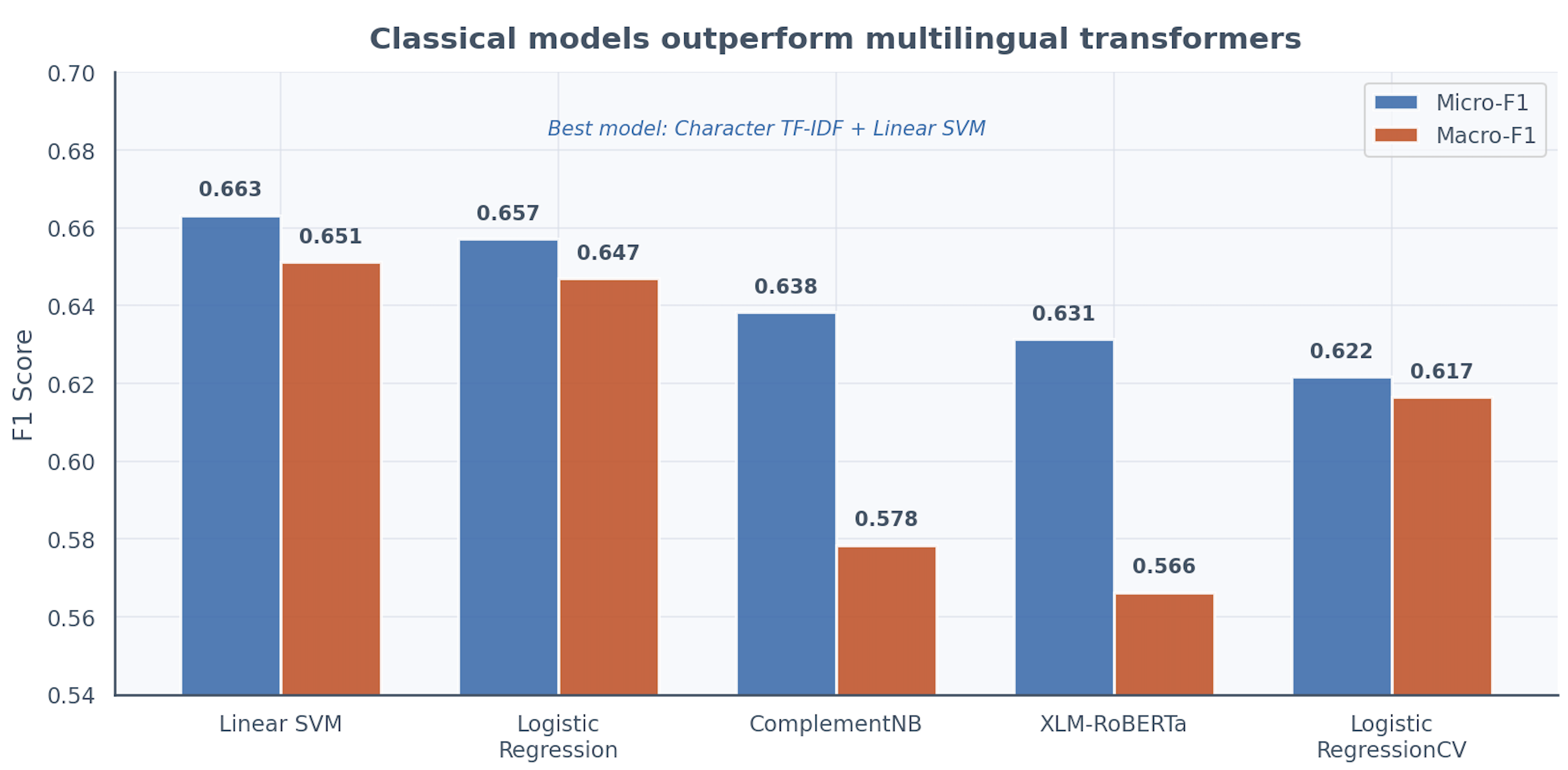}
\caption{Model comparison: Micro-F1 and Macro-F1 for all systems (tuned thresholds).
  Classical TF-IDF models outperform both transformers on this small corpus.}
\label{fig:modelcomp}
\end{figure}

\paragraph{Key findings.}
\textbf{(1) TF-IDF beats transformers.}
Best model (B) achieves Macro-\fone{} = 0.651, exceeding ParsBERT (0.566)
and XLM-RoBERTa (0.588).
With 515 training instances, transformer fine-tuning risks overfitting.
character $n$-grams directly exploit Dari morphological surface patterns
\cite{Cavnar1994}.

\textbf{(2) Threshold tuning is essential.}
Default threshold gives Micro-\fone{} = 0.554; tuned gives 0.657
(+10.3 pp). This gap dwarfs the difference between competing classifiers.

\textbf{(3) Label frequency predicts F1.}
\textsc{Stress} (F1=0.791), \textsc{EduClose} (0.789), and
\textsc{Uncertain} (0.771) score highest, driven by high support and
distinctive Dari keyword patterns.
\textsc{Health} (0.479) remains hardest: its constituent phenomena
(sleep, illness) surface through highly varied phrasing.

\section{Pioneering the Dari NLP Landscape}
\label{sec:pioneering}

Despite Dari being spoken by over 30 million people \cite{Perry2005},
it occupies Joshi et al.'s \citeyear{Joshi2020} ``neglected'' language tier.
To the best of our knowledge, \afstress{} is the \textbf{first-person self-reported stress narratives in Dari.}

Three pillars:
\textbf{(1) Data void addressed:} prior Dari resources are news wire or
event-reaction corpora \cite{Hussiny2023}, not first-person psychological
disclosure;
\textbf{(2) Gold standard established:} documentation of the Unicode bug,
ZWNJ handling, and threshold tuning constitutes the foundational technical
specification for Dari multi-label NLP;
\textbf{(3) Active frontier opened:} \afstress{} provides citable datasets,
evaluation protocols, and a psychological framework enabling uptake across NLP \cite{Farahani2021,Conneau2020}, social science \cite{Charlson2019,Pearlin1989}, and gender studies \cite{UNESCO2025}.

\begin{findingbox}
\textbf{Landmark claim:} \afstress{} moves Dari from a neglected
resource category to an active interdisciplinary frontier, providing
enabling infrastructure for culturally-aware stress and well-being NLP for Afghan communities, at precisely the moment when those communities
need it most.
\end{findingbox}

\section{Limitations}
\label{sec:limitations}
Several limitations bear on the generalisability of \afstress{}. First, recruitment was conducted via online dissemination through
existing community networks (e.g., student groups, educational organisations,
and social media platforms). As participation required internet access,
digital literacy, and awareness of the survey channels, the sample is
likely to overrepresent individuals who are more connected, educated,
and socially engaged. Populations in remote or infrastructure-limited
regions, where internet connectivity is unstable, restricted, or unavailable, 
are systematically underrepresented.

Second, Afghanistan’s current digital environment is characterised by
uneven connectivity, frequent network disruptions, and regional disparities
in access to mobile data and electricity. In some provinces, internet access
is intermittent or prohibitively expensive, which constrains participation
in online surveys and contributes to the relatively modest sample size
($N=737$) despite broad dissemination efforts.

Third, the 90\% female composition, while substantively meaningful as a
reflection of the disproportionate impact of the education ban and social
restrictions on women, limits straightforward generalisation to male
populations. Male-specific stress patterns should be interpreted with
caution due to smaller subgroup sample sizes and higher statistical
uncertainty.

Fourth, responses are self-reported and collected through an online survey,
which introduces potential self-selection bias: individuals experiencing
higher levels of stress or stronger motivation to share their experiences
may be more likely to participate. Additionally, responses reflect perceived
and reported experiences rather than clinically validated assessments.

Finally, the corpus captures a single cross-sectional snapshot, precluding
longitudinal analysis of stress trajectories or causal inference. Observed
co-occurrence patterns and cascade structures should therefore be interpreted
as correlational and theory-consistent rather than causal or predictive.
\section{Ethical Considerations}
\label{sec:ethics}

All responses were anonymized; no personally identifiable information was collected,
and free-text responses were redacted for names and locations where necessary.
Participation was voluntary and based on informed consent. Ethics clearance was obtained from the relevant institutional review process [omitted for blind review]. The corpus is released under a restricted
research license prohibiting re-identification and surveillance-related use.
\afstress{} is a survey corpus of self-reported experiences; it does not constitute
a clinical assessment, psychological diagnosis, or medical record
\cite{Bender2021,Kleinman1988}.
\section{Conclusion}
\label{sec:conclusion}

We introduced \afstress{}, the first survey-based multi-label corpus of self-reported stress narratives in Dari: 737 responses, 12 binary labels, mean cardinality 5.54, and a three-layer interdisciplinary analysis framework.

\textbf{Computationally:} character TF-IDF + Linear SVM achieves
Macro-\fone{} = 0.651 with per-label threshold tuning, outperforming
both ParsBERT and XLM-RoBERTa; threshold tuning provides +10.3 pp
Micro-\fone{} and should be standard practice.

\textbf{Socially:} structural stressors (uncertain future 62.6\%,
education closure 60.0\%) exceed emotional responses, demonstrating
structurally-driven stress; gendered patterns confirm the education ban
as a direct psychological harm mechanism
\cite{UNESCO2025}.

\textbf{Psychologically:} Hopelessness--Uncertain~future ($r=0.40$,
$J=0.388$, 329 joint instances) is consistent with population-scale learned helplessness \cite{Abramson1989,Seligman1975}; mean cardinality
5.54 is the computational fingerprint of chronic allostatic load
\cite{McEwen2007}; and the Jaccard similarity network reveals the cascade
Stress $\to$ Anxiety $\to$ Sadness $\to$ Burnout $\to$ Hopelessness,
consistent with cascade theory \cite{Selby2010}.

\afstress{} moves Dari from a neglected resource category to an active
interdisciplinary frontier \cite{Joshi2020}, enabling future work in
low-resource NLP, computational social science, public health, and the
development of culturally-aware stress and well-being NLP tools for Afghan communities.

\section*{Data Availability Statement}

The \afstress{} dataset will be made publicly available upon acceptance of this paper. Prior to release, all records will be de-identified to remove personally identifiable information and to ensure compliance with ethical and data protection requirements.


\end{document}